\documentclass{llncs}

\usepackage[T1]{fontenc}
\usepackage[utf8]{inputenc}
\usepackage{amsmath}
\usepackage{amsfonts}
\usepackage{comment}
\usepackage{xcolor}

\usepackage[numbers]{natbib}

\usepackage{local-macros}

\begin{document}
\title{Differentiable Logics for Neural Network Training and Verification}
\author{Natalia Slusarz \and Ekaterina Komendantskaya \and
	Matthew L. Daggitt  \and Robert Stewart}
\institute{
	Heriot-Watt University, Edinburgh, United Kingdom \\
	\{nds1, ek19, md2006, rs46\}@hw.ac.uk
}
\maketitle

\section{Introduction}
	The rising popularity of neural networks (NNs) in recent years and their increasing prevalence in real-world applications have drawn attention to the importance of their verification. While verification
is known to be computationally
difficult theoretically~\cite{KaBaDiJuKo17Reluplex}, many techniques have been
proposed for solving it in practice~\cite{BaLiJo21}.

It has been observed in the literature that by default neural networks rarely satisfy logical constraints that we want to verify. A good course of action is to train the given NN to satisfy said constraint prior to verifying them~\cite{hu2016harnessing,pmlr-v80-xu18h}. This idea is sometimes referred to as continuous verification~\cite{KKK20,CKDKKAE22}, referring to the loop between training and verification.

Usually training with constraints is implemented by specifying a translation for a given formal logic language into loss functions. These loss functions are then used to train neural networks. Because for training purposes these functions need to be differentiable, these translations are called \emph{differentiable logics} (DL).

This raises several research questions. What kind of differentiable logics are possible? What difference does a specific choice of DL make in the context of continuous verification? What are the desirable criteria for a DL viewed from the point of view of the resulting loss function? In this extended abstract we will discuss and answer these questions.



\section{Differentiable Logic and Constraint Cased Loss Functions}
\label{sec:dl}
We will explain the main idea behind DL by means of concrete examples. Recall that in machine learning loss functions are used during training to calculate the error between the neural network's current output and the desired output. For example cross-entropy, a popular choice of loss function for classification problems, is defined as follows:

\begin{definition}[Cross Entropy Loss Function]
	\label{def:ce}
	Given a data set\\ $\mathcal{D} =  \{(\x_1, \y_1), \ldots , (\x_n, \y_n)\}$ and a function (neural network) ${f: \Real^n \rightarrow [0,1]^m}$, the cross-entropy loss is defined as 
	\begin{equation}\label{eq:ce}
	\losssymbol_{ce}(\x, \y) = - \sum_{i=1}^{m} \y_i \; \log(f(\x)_i)
	\end{equation}
	where $(\x,\y) \in \mathcal{D}$, $\y_i$ is the true probability for class $i$ and $f(\x)_i$ the probability for class $i$ as predicted by $f$ when applied to $\x$.
\end{definition}

 We now define a small formal language $\Phi$: let the terms $t, t' \in T$ be either variables or constants. Atomic propositions (also called atomic formulae) $A, A_i \in \mathcal{A}$ are given by comparisons between terms using binary predicates $\leq, \neq$. Let us also have conjunction $\wedge$ and negation $\neg$. 
	
	\begin{equation*}
		\Phi \ni \phi, \phi_1, \phi_2 = A\ |\ \phi_1 \wedge \phi_2\ |\ \neg \phi
	\end{equation*} 

Let us use this toy syntax to introduce several different DLs that exist in the literature. To do so we will use the notation $\| \cdot \|: \Phi \rightarrow \mathcal{D}$ for some target domain $\mathcal{D}$ to talk about the possible translations. 





\subsubsection{DL2 translation}
	A good first example is a system called DL2 (Deep Learning with Differentiable Logic) \cite{fischer2019dl}. The translation function $ \|\cdot \|_{DL2} : \Phi \rightarrow [0, \infty]$ is defined as follows.
	
	The definition starts with atomic formulae. The loss for $t \leq t'$ is proportional to the difference between terms $t$ and $t'$ and the loss for term inequality is non-zero when the terms are equal: 
	\begin{align*}
		\tDLtwo{t \leq t'}&:=\max (t-t',0)\\
		\tDLtwo{t \neq t'} &:= \xi [t=t']
	\end{align*}
	where $\xi > 0$ is a constant and $[\cdot]$ an indicator function. And conjunction is defined as:
	\begin{equation}
	\tDLtwo{\phi_1 \wedge \phi_2} := \tDLtwo{\phi_1} + \tDLtwo{\phi_2}
	\end{equation}

	We notice that negation is not explicitly defined in DL2 - it is only defined for atomic formulae since they are comparisons. This is partially related to the choice of domain of the function $\|\cdot \|_{DL2}$. In the interval $[0,\infty]$, 0 denotes \textbf{true} and the rest of the interval denotes \textbf{false}. This interpretation does not admit a simple operation for inversion that could give an interpretation for negation. More generally the choice of interpretation range $[0,\infty]$ is motivated by making the resulting function differentiable ``almost everywhere" (this range resembles the famous activation function ReLu) and give an interpretation of the basic predicates $\leq,\ \neq$. 
	
	Note that we can also view the above translation as a function. For example:
		\begin{align*}
	\tDLtwo{\leq} &= \lambda t, t'.\ max (t-t',0)\\
	\tDLtwo{\neq} &= \lambda t, t'.\ \xi [t=t']\\
	\tDLtwo{\leq} &= \lambda \phi_1, \phi_2.\ \tDLtwo{\phi_1} + \tDLtwo{\phi_2}
	\end{align*}

	In the later sections we will sometimes resort to the functional notation for the translation function.




\subsubsection{Fuzzy logic translation}
Fuzzy DL takes a more conceptual approach to the choice of the domain $\mathcal{D}$. Unlike DL2 which had a focus on interpreting comparisons between terms, fuzzy DL starts with the domain intrinsic to fuzzy logic and develops the DL translation from there~\cite{DBLP:journals/corr/SerafiniG16,DBLP:journals/corr/abs-2006-03472,VANKRIEKEN2022103602}.

	Using our example we look at one implementation of conjunction in fuzzy logic  based on Gödel's t-norm (t-norm, or a triangular norm, is a kind of binary operation used, among others, in fuzzy logic~\cite{cintula2011handbook}). Let us denote this translation as ${\tfuzzy{\cdot}:\Phi\rightarrow[0,1]}$. This time for the base case we assume that atomic formulae are mapped to $[0,1]$ by some oracle. Then:
	
	\begin{equation}
		\tfuzzy{\phi_1 \wedge \phi_2} := \min (\tfuzzy{\phi_1}, \tfuzzy{\phi_2})\\
	\end{equation}

	This time we are given a translation for negation which is

	\begin{equation*}
	\tfuzzy {\neg \phi} := 1-\tfuzzy{\phi}
	\end{equation*}
	

Drawing from the long tradition of fuzzy logic research, \citet{VANKRIEKEN2022103602} survey several other choices of translation for conjunction in fuzzy logic such as Łukasiewicz:
\begin{equation*}
	\tlukasiewicz{\phi_1 \wedge \phi_2} = \max(\tlukasiewicz{\phi_1}+\tlukasiewicz{\phi_2}-1,0))
\end{equation*} 
Yager:
\begin{equation*}
	\tyager{\phi_1 \wedge \phi_2} = \max (1 - ((1-\tyager{\phi_1})^p + (1-\tyager{\phi_2})^p)^{1/p},0),\:\: \text{for}\ p\geq 1
\end{equation*} 
or product based:
\begin{equation*}
	\tproduct{\phi_1 \wedge \phi_2} = \tproduct{\phi_1}\cdot\tproduct{\phi_2}
\end{equation*}
to name a few (see \cite{VANKRIEKEN2022103602} for full survey). All of these logics agree on interpretation of negation.

\subsubsection{Signal Temporal Logic translation}
A different approach by \citet{varnai} was suggested for Signal Temporal Logic (STL) which we adapt to our example language. This paper suggests that the design of DLs should focus on specific properties of the loss functions they give rise to. 

	Let us denote the new translation $\tSTL{\cdot}: \Phi \rightarrow \Real$. \citet{varnai} start with a list of desirable properties of loss functions and create the translation around it. In all of the following we will use $\conjM(A_1, ..., A_M)$ as a notation for conjunction of exactly $ M $ conjuncts. Again the authors assume an oracle (map of atomic formulae to $\Real$) for the base case. 
	
	We assume a constant $\nu \in \Real^+$ and we introduce the following notation:
	\begin{equation*}
		A_{\min} = \min (\|A_1\|_{S}, \ldots, \|A_M\|_{S})
	\end{equation*}
	and
	\begin{equation*}
\tilde{A_i} = \dfrac{\|A_i\|_{S} - A_{\min}}{A_{\min}} 
	\end{equation*}
	Then the translation is defined as:
	\begin{equation}
	\label{eq:stl}
	\|\conjM(A_1,\ldots,A_M)\|_{S}  = 
	\begin{cases}
	\dfrac{\sum_i A_{\min} e^{\tilde{A_i}} e^{\nu \tilde{A_i}}}{\sum_i e^{\nu \tilde{A_i}}} & \text{if}\ A_{\min} < 0 \\
	\dfrac{\sum_i A_{\min} e^{-\nu \tilde{A_i}}}{\sum_i e^{-\nu \tilde{A_i}}} & \text{if}\ A_{\min} > 0 \\
	0 & \text{if}\ A_{\min} = 0 \\
	\end{cases}
	\end{equation}
This translation proposes an elegant notion of negation:
\begin{equation*}
\|\neg \phi\|_{S} = - \|\phi\|_{S}
\end{equation*}

\subsubsection{Use in training}
To use any of these functions in NN training we augment the standard loss function. For example given cross-entropy loss (see Definition \ref{def:ce}), a constraint $ \phi $ and a translation $\tempty{\cdot}$ we would have an augmented loss function:

\begin{equation}
\label{eq:aug_loss}
	\losssymbol_A^{\phi}(\x, \y) = \alpha \losssymbol_{CE}(\x, \y) + \beta \tempty{\phi}
\end{equation}

where $\alpha, \beta \in [0,1]$ are constants.



By looking at these key choices made in literature so far we can see that the choice of a DL involves four major decisions:

\begin{itemize}
	\item \textbf{Domain of interpretation.} We have seen the choices vary between $[0, \infty]$, $[0,1]$ and $[-\infty, +\infty]$.
	\item \textbf{Expressiveness.} Choice of which connectives will be included in the translation, which determines the expressiveness of the language.
	\item \textbf{Interpreting logical connectives.} Although negation is partially determined by the choice of domain, the choice of conjunction seems to be a largely independent decision as evidenced by the presented examples.
	\item \textbf{Interpretation of atomic formulae.} DL2 proposes a concrete approach and some papers leave the definition abstract.
\end{itemize}
In context of continuous verification these choices determine how the translation is implemented in practice. 

\section{Property Based Approach}
\label{sec:properties}

We now consider the mathematical properties of the resulting loss functions. Generally, machine learning research suggests a choice of loss functions for NNs: cross-entropy loss, hinge loss, mean squared error loss, etc. \cite{wang2022comprehensive}. In this community there is some consensus on the desirable properties of loss functions - convexity or continuity are widely considered desirable \cite{klebanov2009robust}. But as we are now focusing on developing methods for logic-driven loss functions, \citet{varnai} also suggest additional desirable properties which we will discuss next. Following from the previous section we continue to assume an oracle mapping of all atomic formulae $\|\cdot \|: \mathcal{A} \rightarrow \mathcal{D}$. 


Soundness relates to the logical satisfaction of the formula. We assume that the language $ \Phi $ has some interpretation $\mathcal{I}$ of its formulae to the set \{\textbf{true, false}\}. As discussed in Section \ref{sec:dl}, the domain in each of the translations is different and that heavily influences what \emph{soundness} will be defined as. Given a domain $\mathcal{D}$ we must divide it into the parts that map to \textbf{true} and \textbf{false}. Let  us denote the part that maps to \textbf{true} as $\mathcal{D}_{true}$.

Let us now define the soundness abstractly.

\begin{definition}[Soundness]
		 A DL is sound for some interpretation $\mathcal{I}$ if for any constraint $\phi$, $\|\phi\| \in \mathcal{D}_{true}$ if and only if constraint $\phi$ is true in interpretation $\mathcal{I}$, denoted as $ \mathcal{I}(\phi) = \textbf{true} $.
\end{definition}

Let us compare the soundness for the specific translations starting from DL2: $\tDLtwo{\cdot}:\Phi \rightarrow [0, \infty]$:

\begin{equation}
	\mathcal{I}(\phi) = \textbf{true} \Leftrightarrow \tDLtwo{\phi} = 0
\end{equation}

For the fuzzy logic we have: $\tfuzzy{\cdot}:\Phi \rightarrow [0,1]$. This is a more intriguing problem as we can no longer assume $ \phi $ is interpreted on boolean values. There is now a choice of splitting the domain to adhere to boolean values as we have used above or using fuzzy logic to express the constraints. In this extended abstract we will use the absolute truth from fuzzy logic instead:

\begin{equation}
\mathcal{I}(\phi) = 1 \Leftrightarrow \tfuzzy{\phi} = 1
\end{equation}

And lastly for the STL translation $\tSTL{\cdot}: \Phi \rightarrow \Real$, we have:

\begin{equation}
\mathcal{I}(\phi) = \textbf{true} \Leftrightarrow \tSTL{\phi} > 0
\end{equation}
As we can see the soundness here is different from how loss functions are usually constructed - they usually have values greater or equal to zero - this choice however is connected to the shadow-lifting property which we discuss later.

These are all significantly different soundness statements which poses a question of how much does the choice of logic impact the mathematical properties of the function that we get from the translation - or conversely whether some properties we may want our function to have can limit the choice of logic we can use.




\begin{definition}[Commutativity, Idempotence and Associativity]
	The \textbf{and} operator $\conjM$ is commutative if for any permutation $\pi$ of the integers $i \in {1, ..., M}  $
	\begin{equation*}
	\|\conjM (A_1, ..., A_M)\| = \|\conjM (A_{k_{\pi(1)}}, ..., A_{k_{\pi(M)}})\|
	\end{equation*}
	it is idempotent if 
	\begin{equation*}
	\|\conjM(A, ..., A)\| = \|A\|
	\end{equation*}
	and associative if
	\begin{equation*}
		\tempty{\conj{2}(\conj{2}(A_1, A_2), A_3)} = \tempty{\conj{2}(A_1, \conj{2}(A_2, A_3))}
	\end{equation*}
	
\end{definition}

Commutativity, idempotence and associativity are identities that make it far easier to use the translation, as changes in the order of elements in conjunction will not affect the resulting loss function. It is also important to note that associativity is not a part of original set of desirable properties as listed by \citet{varnai} and is not satisfied by the translation $\tSTL{\cdot}$ - which is the reason why they define conjunction as an n-ary rather than a binary operator.

Before defining the next property we should also take a look at the notion of \textit{gradient}, which for a differentiable function $f$ is a vector of its partial derivatives at a given point.

\begin{definition}[Weak smoothness]
	The $\|\conjM\|$ is \textit{weakly smooth} if it is continuous everywhere and its gradient is continuous at all points $\|A_i\|$ with ${i \in \{1, ..., M\}}$ for which no two indices $ i \neq j $ satisfy
	 $ \|A_i\| = \|A_j\| = \min(\|A_1\|, ..., \|A_M\|) $.
\end{definition}

In more informal terms we require smoothness at points where there is a unique minimal term. The specific definition of weak-smoothness holds in particular for points where the metric switches signs (and therefore, by definition, its truth value). 

\begin{definition}[Min-max boundedness]
	The operator $\|\conjM\|$ is \textit{min-max bounded} if
	\begin{equation*}
	\min(\|A_1\|, ..., \|A_M\|) \leq \|\conjM(A_1, ..., A_M)\| \leq \max(\|A_1\|, ..., \|A_M\|)
	\end{equation*}
\end{definition}

The min-max boundedness ensures closure of the translation with respect to the domain.

\begin{definition}[Scale invariance]
	The interpretation of $\conjM$ is said to be \textit{scale-invariant} if, for any $ \alpha \leq 0$ with $\alpha \in \Real$
	\begin{equation*}
	\alpha \|\conjM (A_1, ..., A_M)\| = \tbig{\conjM} (\alpha \|A_1\|, ..., \alpha \|A_M\|)
	\end{equation*}
\end{definition}

With \textbf{scale-invariance} we can be sure that the metric will behave in a similar manner regardless of the magnitude (in case it was unknown).

Now  ``shadow-lifting'' is a property original to \cite{varnai}. Its motivation is to encourage gradual improvement when training the neural network even when the property is not yet satisfied. In other words if one part of the conjunction increases the value for the translation of entire conjunction should increase as well. 

\begin{definition}[Shadow-lifting property]
	Let $ \|\conjM\| $ satisfy the \textit{shadow-lifting property} if, $\forall i. \|A_i\| \neq 0$:
	
	\begin{equation*}
	\left. \dfrac{\partial \|\conjM(A_1, ..., A_M)\|}{\partial \|A_i\|}\right\rvert_{A_1, ..., A_M} >0
	\end{equation*}
	where $ \partial $ denotes partial differentiation.
\end{definition}


It is this property that motivated the translation of conjunction $\tSTL{\cdot}$ described in Equation~\ref{eq:stl}. It is different to the classical notion of conjunction which, if not satisfied, does not preserve information when it comes to value of individual conjuncts or their number. It is interesting to consider whether it would be more useful to use a logic with two variants of each connective - keeping the classical FOL connectives as well as ones adhering to the shadow-lifting property.

 There is also the issue of domain which we've touched on before. This translation allows the values of the translation to range across entire real domain ($\Real$) - defining \textit{true} as greater then zero and allowing us to define negation by simply flipping the sign. While this approach was viable for reinforcement learning that this metric was designed for, it creates a problem of compatibility when it comes to training neural networks. Typically a constraint loss function that is generated is not used on its own - but in combination with a more classical one such as cross-entropy loss (see Equation \ref{eq:aug_loss})- and would have to be scaled appropriately if we do not want it to imbalance the training.

This poses a more general question of how the choice of properties that we deem as desirable determines the choice of logical syntax that we can use.

Let us see how the DLs that we have introduced compare when it comes to satisfying these properties.

\begin{table}
\begin{center}
		\caption{Property comparison between different DL translations of conjunction. Properties which have been stated in the relevant paper or proven in other works have relevant citations.}
	\label{tab:properties}
	
	\begin{tabular}{|p{0.2\linewidth}|c|c|c|c|c|c|}
		\hline 
		 & $\tDLtwo{ \phi_1 \wedge \phi_2 }$ & $ \tfuzzy{ \phi_1 \wedge \phi_2 } $ &$\tlukasiewicz{\phi_1 \wedge \phi_2}$&$ \tyager{\phi_1 \wedge \phi_2} $&$ \tproduct{\phi_1 \wedge \phi_2} $&$ \tSTL{ \phi_1 \wedge \phi_2 } $\\ 
		\hline 
		& $ [0, \infty]$ &$ [0,1]$&$ [0,1]$&$[0,1]$&$[0,1]$& $ [-\infty, \infty] $\\ 
		\hline \hline
		\textbf{Properties:} & &&&&&\\
		\hline \hline

		\hline 
		Idempotent & no & yes \cite{VANKRIEKEN2022103602} &no \cite{cintula2011handbook}&no \cite{klement2004triangular}&no \cite{cintula2011handbook}& yes \cite{varnai}\\
		\hline 
		Commutative & yes & yes \cite{VANKRIEKEN2022103602} &yes \cite{VANKRIEKEN2022103602}&yes \cite{VANKRIEKEN2022103602}&yes \cite{VANKRIEKEN2022103602}& yes \cite{varnai}\\
		\hline 
		Shadow-lifting  & yes & no &no&no&yes& yes \cite{varnai}\\
		\hline 
		Min-max boundedness & no & yes &no&no&yes& yes \cite{varnai}\\
		\hline 
		Scale invariance & yes & yes &no&no&no& yes \cite{varnai}\\
		
		\hline
		Associativity &yes &yes \cite{VANKRIEKEN2022103602}&yes \cite{VANKRIEKEN2022103602}&yes \cite{VANKRIEKEN2022103602}&yes \cite{VANKRIEKEN2022103602}&no \cite{varnai}\\
		\hline
		
	\end{tabular}
	\\

\end{center}
\end{table}

 
 

 Table \ref{tab:properties} summarises results already established in literature as well as provides new results. Let us briefly describe the reasoning behind the table entries that gave new results. It is important to note that the properties take into account the domains of each translation. 

Starting with shadow-lifting, G\"{o}del, Łukasiewicz and Yager t-norms all do not have that property. These translations involve minimum or maximum which both do not preserve shadow-lifting. In all of these cases there are cases at which change of value of one of the conjuncts will not change the value of $\min\backslash\max$. But the property holds for both DL2 and product translations for most properties due to them being defined simply by addition and multiplication respectively.

With min-max boundedness the reasoning is different for each translation. Interestingly here the product translation is bounded due to the domain being $[0,1] $. By definition of minimum the G\"{o}del based DL is also bounded while both Łukasiewicz and Yager can return values greater then the largest value of a conjunct.

For the case of scale invariance DL2, G\"{o}del and product entries are trivial. Both Łukasiewicz and Yager inspired DLs are not scale invariant due to the terms inside $\max$ containing addition or subtraction of constants not dependant on the individual conjuncts.

All of the fuzzy logic translations are associative as they are based on t-norms which are associative by definition~\cite{VANKRIEKEN2022103602} and DL2 is associative as it is defined by addition. The STL based metric is the only one which is not associative - associativity together with idempotence would prevent shadow-lifting which was deemed more desirable~\cite{varnai}.

We can see that none of the translations have all of these properties. This shows that the property oriented approach to finding a translation is non-trivial and the choice of a DL heavily influences the properties of the resulting loss function.

\section{Conclusions and future work}
\label{sec:design_space}
\subsection{Conclusions}

In this extended abstract we answered the questions posed in the abstract. Firstly, we presented a uniform way of defining a translation from logic syntax to a differentiable loss function. This has allowed us to compare them in terms of the mathematical properties of their translation, providing an overview of the current state of the art when it comes to DLs. This in turn allowed us to reason about the design space of future DLs - the properties we may want them to have, the choice of domain, choice of logic etc. 
This is the first step to providing a comprehensive guide to creating new DLs while being conscious of the consequences certain design decisions can bring for continuous verification of NNs.

\subsection{Future work and design space}

We have briefly mentioned at the end of Section \ref{sec:dl} some of the design choices that one has to face when choosing and designing a DL, as well as its interpretation to a loss function for training neural networks. When we try to compare the different DLs we can group the trade-offs in a few categories:

\subsubsection{Expressive power of DLs}
\textit{Expressive power} is a broad category that we have mentioned in Section \ref{sec:dl}. It is impacted by the choice of logic and its domain of interpretation as those two things can limit, for example, the choice of defined connectives as can be seen in the following example of DL2. 

The lack of negation in the defined syntax of DL2 is a direct consequence of the domain ($\mathcal{D} = [0,\infty])$. The only reason the translation does not lead to a significant loss of expressiveness is due to the explicit translation of all atomic formulae and their negations - this way one can ``push'' the negation inwards in any constraint. In case all other translations discussed before, which all leave the interpretation of atomic formulae to an oracle, it would only work if we added an assumption that said oracle also interprets negation of every atomic formula.

Meanwhile for fuzzy logic DLs there is a well defined domain, however we encounter an issue when trying to split it to assign boolean values for the purposes of translation. We need to have a fully \textbf{true} state - state when the constraint is fully satisfied. This creates a choice of how the domain should be split, which heavily impacts the expressiveness of the DL. 
\subsubsection{Mathematical properties of DLs}
We have already discussed how different DLs compare in terms of \textit{mathematical properties} (see Section \ref{sec:properties}). Some of these properties are matters of convenience - associativity and commutativity for example ensure that order of elements in conjunction will not affect the resulting loss function. Others, such as shadow-lifting, change the way the loss function will penalise NNs behaviour and therefore the way it is trained.

 This comparison comes down to interpretation of logical connectives. We have presented multiple interpretations of conjunction which, while of course influenced by the domain, give a lot of freedom in their design. Fuzzy logic based DLs are a great example of this considering they resulted in multiple translations with the same domain and syntax and different mathematical properties and behaviour (see Section \ref{sec:properties} for a detailed discussion). 



While both the syntax being translated and the mapping of values are dependant on properties the properties themselves are not immutable. This leads to a question of what other properties would we want to add to the list of ones we consider desirable and which should be prioritised - for example among the current properties it is impossible to satisfy idempotence, associativity and shadow-lifting simultaneously. It is also important to mention that in this study we have we omitted completeness in this study, yet it deserves further investigation in the future.
\subsubsection{Training Efficacy and Efficiency}
From an implementation perspective there is also the question of \textit{efficiency} which we have not discussed before - for example we may prefer to avoid non-transcendental functions or case splits - both of which were present in some translations - as they make it far more costly to train the network.

\subsubsection{}An immediate plan for the future involves developing a new translation into a differentiable loss function, taking some inspiration from the property-driven approach. In this talk we will discuss how the different translations presented compare when it comes to performance - based on the results we will be able to draw more conclusions about the design space for DLs in future work.




\newpage
\bibliographystyle{plainnat}

\bibliography{Natalia}
\end{document}